\documentclass{article}
\usepackage{ijcai19}

\usepackage{amssymb} 

\usepackage{times}
\usepackage{soul}
\usepackage{url}
\usepackage[hidelinks]{hyperref}
\usepackage[utf8]{inputenc}
\usepackage[small]{caption}
\usepackage{graphicx}
\usepackage{amsmath}
\usepackage{booktabs}
\usepackage{xspace}
\usepackage{xcolor} 
\definecolor{ForestGreen}{rgb}{0.13, 0.55, 0.13}
\newcommand{\eegmamba}{EEG-SSM\xspace}

\title{\eegmamba: Leveraging State-Space Model for Dementia Detection}

\author{
Xuan-The Tran$^1$\and
Linh Le$^1$\and
Quoc Toan Nguyen$^1$\And
Thomas Do$^1$\\
Chin-Teng Lin$^1$\footnote{Contact Author}\\
\affiliations
$^1$University of Technology Sydney (UTS), Faculty of Engineering and Information Technology, GrapheneX-UTS Human-centric AI centre, Sydney, New South Wales, 2007, Australia\\
}

\begin{document}

\maketitle

\begin{abstract}
State-space models (SSMs) have garnered attention for effectively processing long data sequences, reducing the need to segment time series into shorter intervals for model training and inference. Traditionally, SSMs capture only the temporal dynamics of time series data, omitting the equally critical spectral features. This study introduces \eegmamba, a novel state-space model-based approach for dementia classification using EEG data. Our model features two primary innovations: \eegmamba\ temporal and \eegmamba\ spectral components. The temporal component is designed to efficiently process EEG sequences of varying lengths, while the spectral component enhances the model by integrating frequency-domain information from EEG signals. The synergy of these components allows \eegmamba\ to adeptly manage the complexities of multivariate EEG data, significantly improving accuracy and stability across different temporal resolutions. Demonstrating a remarkable 91.0\% accuracy in classifying Healthy Control (HC), Frontotemporal Dementia (FTD), and Alzheimer's Disease (AD) groups, \eegmamba\ outperforms existing models on the same dataset. The development of \eegmamba\ represents an improvement in the use of state-space models for screening dementia, offering more precise and cost-effective tools for clinical neuroscience.
\end{abstract}

\section{Introduction}
Dementia, a cognitive decline that impairs memory and functional abilities, affects approximately 45 million people globally \cite{nichols2019global}, with numbers expected to increase due to longer lifespans and an aging population. Alzheimer's disease (AD), the most common form of dementia, accounts for 60-80\% of cases \cite{alzheimer20182018}, affecting 1 in 9 individuals over 65 \cite{rajan2021population}. It is a progressive neurological disorder characterized by beta-amyloid accumulation and tau tangles, leading to neuronal death, brain atrophy, and symptoms like memory loss, confusion, and behavioural changes. Frontotemporal degeneration (FTD), another form of dementia, leads to social, behavioural, and motor deficits due to the deterioration of frontal and temporal lobes. FTD typically affects individuals aged 45-60 and represents a significant portion of dementia cases, especially in those under 65 \cite{hogan2016prevalence}. These dementia present significant healthcare challenges with substantial impacts on people with dementia (PwD) \cite{beerens2014quality} and society, underscoring the need for advancements in accurate screening methods.

The standard screening approach for dementia typically involves cognitive tests like the Mini-Mental State Examination (MMSE) \cite{folstein1975practical}, the Montreal Cognitive Assessment (MoCA) \cite{julayanont2017montreal}, the Mini-Cog test \cite{borson2000mini}, and the Addenbrooke's Cognitive Examination III (ACE-III) \cite{bruno2019addenbrooke}. These tests are advantageous due to their simplicity and quick administration. However, their effectiveness can be compromised by factors such as the patient's educational background, emotional state, and test reiteration. Neuroimaging techniques like MRI and PET scans offer greater detecting accuracy but are hampered by high costs and the necessity for extensive expert analysis \cite{sun2024ensemble}. Electroencephalography (EEG) serves as a practical alternative, being cost-effective, non-invasive, and highly portable while providing direct, high-resolution monitoring of brain activity. The development of AI-based detecting models using EEG data can potentially yield more accurate detection and reduce the burden on healthcare professionals.

The quest for effective AI-based detection models for brain disorders has led to the exploration of various machine-learning techniques using EEG data. Traditional machine learning models typically utilize temporal, spectral, and spatial features extracted from EEG signals for classification tasks. A notable study by McBride et al. \cite{mcbride2014spectral} employed a support vector machine (SVM) to distinguish between healthy controls (HC), mild cognitive impairment (MCI), and PwD, achieving an accuracy rate of 85.4\%. Deep learning approaches have also been investigated for this purpose. Ieracitano et al. \cite{ieracitano2019convolutional} used a multilayer perceptron (MLP) model incorporating wavelet EEG features to classify the same groups, achieving an 89.2\% accuracy. Another innovative study applied a Boltzmann machine to spectral images from EEG channels, achieving an impressive accuracy of 95\%. Ensemble learning methods, which combine multiple machine learning models, have shown promising results. For instance, Sun et al. \cite{sun2024ensemble} leveraged an ensemble approach using Adaboost to differentiate between HC, mild cognitive impairment (MCI), and dementia groups with an accuracy of 93.3\% across a dataset including 75 HC, 99 with MCI, and 78 with dementia. Kim et al. \cite{kim2023deep} extended this approach with an ensemble deep learning model validated on a substantial dataset of 1379 EEG recordings from 1155 PwD. 

In the dataset leveraged for our research, Miltiadous et al. reported an accuracy of 83.28\% for Alzheimer's disease versus Control (AD-CN) and 74.96\% for Frontotemporal Dementia versus Control (FTD-CN) using a Convolution-Transformer Architecture known as DICE-Net \cite{miltiadous2023dice}. A separate method by Chen et al., which integrated Convolutional Neural Networks (CNNs) with Visual Transformers (ViTs), attained an accuracy of 76.37\% \cite{chen2023multi}. Our results will be juxtaposed with these findings for a comprehensive comparison.

The burgeoning computational power and the advent of transformer-based deep learning methods have ushered in new opportunities for EEG data analysis. With their parallel computation abilities and self-attention mechanisms, transformers are well-suited for capturing the intricate temporal patterns within EEG signals that extend over long durations \cite{vaswani2017attention}. However, despite their advantages, transformers are not without their challenges. They often necessitate substantial computational resources, and their performance tends to wane as sequence lengths grow \cite{mamba}. This is a significant concern for EEG data characterized by complex and prolonged temporal dynamics.
To surmount the inherent limitations of transformers, a novel approach known as the State Space Models with Selective Scan, or Mamba models, has been introduced \cite{mamba}. These models strive to synergize the sequential processing strength inherent in traditional state space models (SSM) \cite{hamilton1994state} with the parallel processing and long-term memory retention proficiencies of transformers. Mamba models are thus poised to capitalize on the best of both worlds, offering a promising new direction for enhancing EEG data analysis and expanding the potential for AI-based screening in neurology. Several variations of the Mamba model have been successfully applied in both the vision ~\cite{zhu2024vision,liu2024vmamba,ma2024u,liu2024swin} and medical fields \cite{guo2024mambamorph}, yielding promising results.

In this study, we propose an improved Mamba model by adapting it to varying temporal resolutions of multivariate EEG data. Our adaptation, termed  \eegmamba$_{\text{temporal}}$, demonstrates the model’s versatility in handling different EEG segment sequence lengths. Furthermore, we have developed \eegmamba$_{\text{spectral}}$, a novel approach that enables the Mamba model to integrate spectral information from EEG signals. By merging \eegmamba$_{\text{temporal}}$ and \eegmamba$_{\text{spectral}}$, we formulate \eegmamba$_{\text{combined}}$, a hybrid model that leverages the strengths of both temporal and spectral domains to achieve enhanced accuracy and stability, even as EEG segment length increases. Crucially, \eegmamba$_{\text{spectral}}$ addresses individual variability in EEG signals by determining the optimally weighted contributions of various EEG frequency bands for each subject's segment data. Our integrated approach not only showcases the adaptability and efficiency of \eegmamba models but also lays the groundwork for developing more precise and computationally economical detection tools in clinical neuroscience.

\section{Method}
\begin{figure*}[h]
    \centering
    \includegraphics[width=1\textwidth]{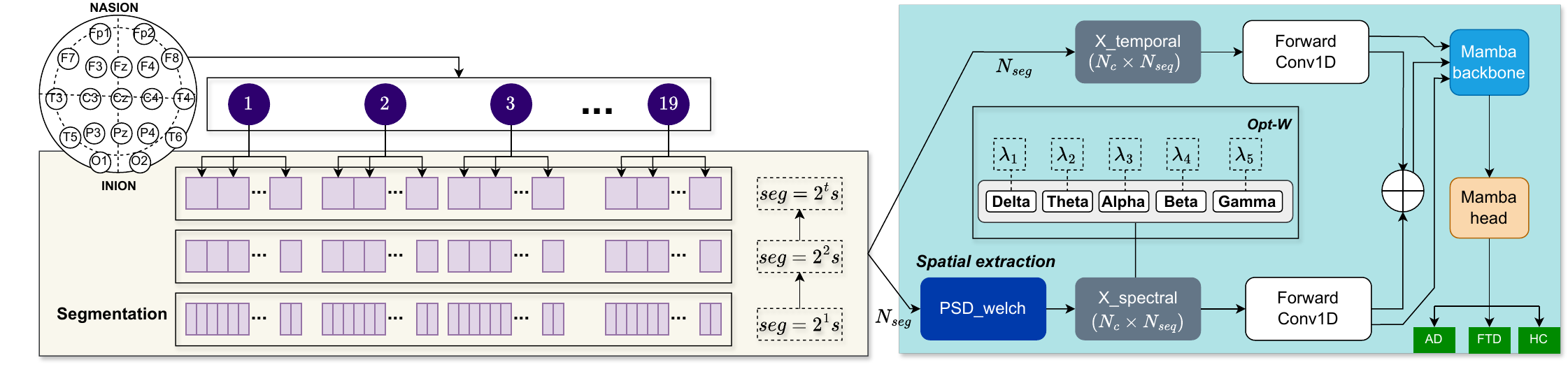}
    \caption{Overview of our proposed framework.}
    \label{our-framework}
\end{figure*}
Our \eegmamba model aims to harness the strengths of SSM in conjunction with multivariate EEG segmentation, as detailed in Figure \ref{our-framework}. This section begins with an introduction to the key principles underpinning SSMs. It then delves into the technique used to divide EEG data into multivariate segments, each defined by specific feature categories: temporal and spectral features (frequency). The section concluded by presenting \textit{a novel optimization strategy (Opt-W)} designed to efficiently encode EEG wavelet bands, showcasing the innovative integration of SSM with EEG data analysis. 
\subsection{Preliminaries - SSM}
The SSM functions by mapping the input vector \( x \in \mathbb{R}^{1 \times D} \) to a higher-dimensional latent space \( h \in \mathbb{R}^{1 \times N} \) (where \( N > D \)), subsequently computing a mapping from \( h \) to the output \( y \in \mathbb{R}^{1 \times D} \). The following set of equations formalizes the SSM:
\begin{equation} 
h'(t) = Ah(t) + Bx(t) \label{mamba-input}
\end{equation}
\begin{equation}
    y(t) = Ch(t) + Dx(t)
\end{equation}
where \( A \in \mathbb{R}^{N \times N} \), \( B \in \mathbb{R}^{N \times 1} \), \( C \in \mathbb{R}^{1 \times N} \), and \( D \in \mathbb{R}^{1 \times 1} \) represent the learned parameter matrices of the model.
Equations 1 and 2, initially in continuous form, are not directly applicable for deep learning tasks. To adapt these for use, they were converted to a discrete form using the Zero-Order Hold (ZOH) method \cite{pohlmann2000principles}. Following this transformation, a global convolution operation was applied to the discretized form, rendering it suitable for input into the Mamba Encoder.
\begin{equation}
K = \begin{pmatrix}
C\tilde{B} & CA\tilde{B} & \cdots & CA^{M-1}\tilde{B}
\end{pmatrix},
\end{equation}
\begin{equation}
y = x * K, 
\end{equation}
where \( M \) signifies the length of the input sequence \( x \), and \( K \) is an \( \mathbb{R}^M \) - dimensional structured convolutional kernel.



\subsection{EEG Preprocess}
\subsubsection{Segmentation}
The Mamba architecture, originally tailored for 1-D sequences, has been adapted in this work for multivariate EEG data to encapsulate all EEG feature dimensions, as depicted in Figure \ref{our-framework}. The upper module of the architecture processes segmented multivariate EEG data, preparing it for the \eegmamba encoder block. 
A key attribute of the Mamba model is its proficiency in managing long sequence data, rendering it particularly apt for resting-state EEG analysis. This attribute allows for EEG data segmentation into various lengths with minimal data information loss. 
\paragraph{Temporal EEG:}To process the multivariate EEG data into blocks suitable for encoder networks, we initiate by converting the 3-dimensional EEG data, denoted as $t\in{\mathbb{R}}^{N_{seg}\times{N_c}\times{N_{seq}}}$, into flattened 2-dimensional patches represented by $x_p\in\mathbb{R}^{{N_c}\times{N_{seq}}}$. Here, $N_{seg}$ signifies the number of segments, $N_c$ stands for the number of channels, and $N_{seq}$ denotes the segment length corresponding to the number of time points within a single EEG segment. For instance, a segment with a length of 2 seconds could encompass a sequence ranging from time frame = 0 to time frame = 2s. Subsequently, these 2-D patches $x_p$ are linearly projected onto a vector of size $M$, which facilitates the extraction of temporal features from the EEG data follows:
    $T_0=[t^1_p{S};t^2_p{S};...;t^J_p{S}]$ where $t^j_p$ is the \textit{j-th} patch of the original EEG sequence, $S\in\mathbb{R}^{{N_c}\times{N_{seq}}}$.
The \eegmamba encoder is used in individual segments, extracting representative EEG features for each segment. Given that Mamba operates on 1D sequences, the temporal domain data, \( X_{\text{temporal}} \), retaining the original shape of the EEG segment input, transforms a Forward Conv1D layer into a 1D array as follows:
\begin{equation}
    x(t)=[Conv_{1d}(t^1_p{S});...;Conv_{1d}(t^J_p{S})] \label{temporal-features}
\end{equation}

\paragraph{Spectral EEG}
\begin{itemize}
    \item Wavelet Filter: We employ a Wavelet Filter \cite{xu1994wavelet}, denoted as $\Omega$, to process EEG data for feature extraction. Specifically, we focus on segmenting the EEG signals into distinct frequency bands, each associated with different cognitive states and activities. These bands include the Delta band (0.5– 4Hz), Theta band (4– 8Hz), Alpha band (8–12 Hz), Beta band (12–20 Hz), and Gamma band (20–40 Hz), which are utilized to feature the EEG data for subsequent analysis \cite{sun2024ensemble}.
    \item In the spectral domain, Power Spectral Density (PSD) features are computed using Welch's method \cite{welch1967use} to obtain \( X_{\text{spectral}} \) before its transformation to a 1D array by an additional Forward Conv1D layer follows:
\begin{equation}
    x(t)=[Conv_{1d}(\Omega(t^1_p{X}));...;Conv_{1d}(\Omega(t^J_p{X}))] \label{spectral-features}
\end{equation}
\end{itemize}
\paragraph{Mamba Input}
In our methodology, the Mamba system is designed with three distinct inputs $x(t)$ (see Equation \ref{mamba-input}) : one for purely temporal features (Equation \ref{temporal-features}), another for exclusively spectral features Equation \ref{spectral-features}), and a third that concatenates both temporal and spectral features, allowing for a comprehensive analysis through separate yet complementary data feature types (see Figure \ref{our-framework}).
\subsection{EEG wavelet optimization for spectral features (Opt-W)}
\begin{figure}[!ht]
    \centering
    \includegraphics[width=0.3\textwidth]{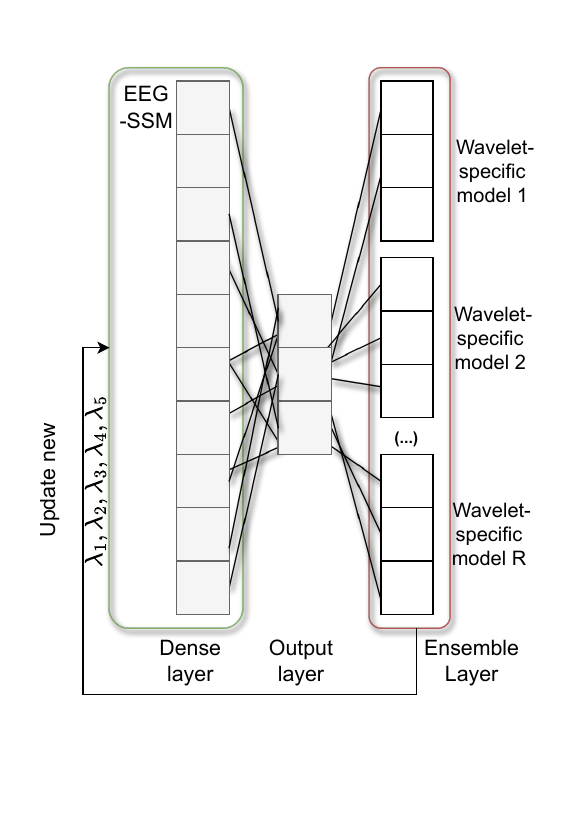}
    \caption{Bottleneck structure for Adaptive EEG Wavelet-Specific Weight Module with three classes and R wavelet-specific models.}
    \label{eeg_5bands}
\end{figure}
Previous research ~\cite{swarnalatha2023greedy,alsharabi2022eeg,sedghizadeh2022network} in the field of cognitive impairment and dementia detection has been limited by an inability to automatically optimize the weights among wavelet bands (Delta, Theta, Alpha, Beta, Gamma). The ratio between these bands plays a critical role in identifying various cognitive conditions \cite{chikhi2022eeg}, underscoring the need for a solution capable of dynamically of each wavelet's importance band according to the specific type of dementia and the patient's condition. Implementing such a system would significantly enhance the accuracy and efficacy of classifying cognitive problems. In this segment, we delineate the ensemble layer, an innovative network layer engineered for the efficacious training of deep neural networks by leveraging features from separate wavelet-specific models to automatically update the weights featuring the combination of different wavelet bands (Delta, Theta, Alpha, Beta, Gamma) for spectral features. The foundational idea is intuitive: the ensemble layer assimilates inputs typically designated for the ultimate layer of a deep neural network (such as a softmax layer for classification) and orchestrates a wavelet-specific linkage from this terminal layer to the labels of various wavelet-centric models. This linkage adeptly encapsulates the distinct reliabilities and the biases introduced by specific wavelet bands, transforming the conventional output layer into a pivotal bottleneck layer that is communally utilized by different wavelet-centric models. For illustrative purposes, Figure \ref{eeg_5bands} portrays this bottleneck configuration within a neural network framework tailored for EEG classification challenges, featuring three classes and encompassing R models, inclusive of 5 wavelet-specific models.

The methodology entails employing the labels from a designated wavelet band to disseminate errors throughout the neural network. The ensemble layer meticulously adjusts the gradients emanating from the labels of that particular wavelet band, aligning them with its accuracy through scaling and bias modulation. Consequently, the bottleneck layer of the network aggregates these tailored gradients from the labels of various wavelet-centric models and propels them backward through the network's remaining structure. This orchestration, facilitated by the so-called ``ensemble layer'', empowers the network to mitigate the influence of unreliable labelers and rectify persistent biases in their labels, all within the traditional backpropagation paradigm.

Formally, let $\xi$ symbolize the output of a deep neural network with an arbitrary architecture. For the sake of simplicity, we postulate that $\xi$ emerges from a softmax layer, with $\xi_c$ denoting the probability of classifying the input instance into category $c$. The activation for each wavelet-specific model $r$ within the ensemble layer is defined as $m^r = f_r(\xi)$, where $f_r$ signifies a wavelet-specific function, and the output of the ensemble layer is rendered as the softmax of these activations: $\psi^r_c = \frac{e^{m^r_c}}{\sum_{l=1}^C e^{m^r_l}}$.

The conundrum then lies in formulating the function $f_r(\xi)$. In the experimental section, we explore various alternatives. For classification endeavors, it is plausible to envisage a matrix transformation such that $f_r(\xi) = W^r\xi$, where $W^r$ represents a wavelet-specific matrix. Given a cost function $E(\psi^r, y^r)$ that juxtaposes the expected output of the $r^{th}$ wavelet-specific model with its actual label $y^r$, we can ascertain the gradients $\frac{\Delta E}{\Delta m^r}$ at the activation $m^r$ for each model and backpropagate them to the bottleneck layer, culminating in the equation:

\begin{equation}
    \frac{\Delta E}{\Delta \xi} = \sum_{r=1}^R W^r \frac{\Delta E}{\Delta m^r}
\end{equation}

Hence, the gradient vector at the bottleneck layer amalgamates into a composite of gradients, each weighted according to the labels from the disparate wavelet-specific models. Furthermore, if a wavelet model is predisposed to misclassify class c as class l (indicative of a bias specific to the wavelet model), the matrix $W^r$ can adeptly recalibrate the gradients to accommodate this bias. The weights of the wavelet-specific models, ${W^r}_{r=1}^R$, delineating the transition from the bottleneck layer's output $\xi$ to the wavelet labels ${\psi^r}_{r=1}^R$, can be deduced using conventional stochastic optimization methodologies, such as SGD or Adam. Subsequent to the network's training phase, the ensemble layer can be detached, unveiling the bottleneck layer's output $\xi$, which can subsequently be harnessed for prognostications on novel instances.
In our exploration, the individual sensitivities $\sigma^r$ and specificities $\delta^r$ are equated to the prediction probabilities and the biases emanating from the training model, respectively.
\section{Experiments}
\subsection{Dataset}
In this study, we utilized a recently published EEG resting state dataset from May 2023, comprising data from 88 subjects across three categories: AD, FTD, and HC \cite{miltiadous2023dataset}. This dataset was selected for its comprehensive data length for each subject and the novelty of its application in research. It features a balanced representation of conditions with 36 AD participants, 23 FTD subjects, and 29 healthy HC. Data were collected using 19 EEG channels (Fp1, Fp2, F7, F3, Fz, F4, F8, T3, C3, Cz, C4, T4, T5, P3, Pz, P4, T6, O1, and O2) and 2 reference electrodes, adhering to the international 10–20 system. The recordings offer a resolution of 10 uV/mm at a sampling rate of 500 Hz. Duration varied among groups: AD recordings averaged 13.5 minutes (ranging from 5.1 to 21.3 minutes), FTD recordings 12 minutes (ranging from 7.9 to 16.9 minutes), and HC recordings 13.8 minutes (ranging from 12.5 to 16.5 minutes), totalling 485.5 minutes for AD, 276.5 minutes for FTD, and 402 minutes for HC. Despite the resting state condition with closed eyes, some recordings exhibited eye movement artifacts. To address this, the dataset employed the Independent Component Analysis (ICA) method (specifically, the RunICA algorithm) to transform the 19 EEG signals into 19 ICA components. Artifacts identified as eye or jaw movements were subsequently removed using the "ICLabel" algorithm within the EEGLAB platform \cite{delorme2007enhanced}, ensuring the integrity of the EEG data for analysis.

\subsection{Settings}
We trained our \eegmamba model on the dataset previously referenced, setting the model dimension (d-model) to match the number of EEG channels, which is 19. The model distinguishes between HC, FTD, and AD; thus, the number of classes was set to 3. For the optimization process, we employed cross-entropy loss and the Adam optimizer \cite{kingma2014adam} with a learning rate of 0.001, iterating the training process over 500 epochs to ensure adequate learning and convergence of the model.

In this study, we trained the \eegmamba model using a dataset that includes a balanced division into training, testing, and validation sets, each constituting 60\%, 20\%, and 20\% of the data, respectively. The model's efficacy was assessed against established benchmarks, such as a basic Recurrent Neural Network (RNN) and an EEG-transformer model. Consistent across all models, we adhered to a training regimen of 500 epochs to ensure thorough learning and fair comparison.

We use a simple RNN model \cite{medsker2001recurrent} for this classification task. The model, structured with an RNN layer followed by a fully connected layer, is designed to process sequential EEG signals. Configured with an input size of 19 to match the EEG features, the RNN uses 128 hidden units across two stacked layers to enhance learning depth. The forward pass initiates with a zeroed initial hidden state, ensuring the model adapts to the input sequence dynamically. This architecture culminates in mapping the sequential analysis to a predefined number of classes, optimizing for accuracy and computational efficiency. Key hyperparameters, including a learning rate of 0.001 and training over 500 epochs, are fine-tuned to balance performance and speed, making this model a strategic choice for discerning patterns in EEG data relevant to dementia classification.

We also implemented a patched EEG-transformer model inspired by the approach described in Chen et al. \cite{Chen_2023_CVPR}. The model utilizes a custom 1-D patch embedding method to project multivariate EEG data into a higher-dimensional space, with a time-len= EEG-segment-length, patch-size=4, embed-dim=1024, and in-chans=19. The encoder consists of depth=24 Transformer blocks, each with num-heads=16, systematically refining the embeddings through self-attention mechanisms to capture complex temporal EEG dynamics. 

\subsection{Dementia detection results}
\begin{table*}[!h]
\centering
\caption{Model performance (accuracy $\%$) for resting state neural disorder classification (19 channels, sampling rate 500Hz)}
\label{test_accuracy}
\begin{tabular}{lcccccccccccccc}
\toprule
Model & Params & $2s$ & $4s$ & $8s$ & $16s$ & $32s$ & $64s$ & $128s$ & $256s$ \\
\midrule
RNN & 8.6K & 41.5 & 40.5 & 40.8 & 40.0 & 42.7 & 24.0 & 37.2 & 39.6 \\
EEG-Net & 12K & 50.3 & 55.6 & 51.3 & 53.4 & 55.6 & 51.5 & 48.8 & 47.5 \\
EEG-Transformer & 120K & 70.1 & 65.6 & 58.3 & 47.4 & 38.6 & 34.5 & 33.7 & 30.5 \\\hline
\eegmamba$_{\text{temporal}}$ & 4.4K & 89.1 & 86.5 & 82.3 & 77.4 & 74.5 & 64.1 & 52.5 & 46.8 \\
\eegmamba$_{\text{spectral}}$ & 4.4K & 80.1 & 79.5 & 79.6 & 79.1 & 78.4 & 77.6 & 77.1 & 76.8 \\
\eegmamba & 4.4K & 91.0 & 87.5 & 83.3 & 79.7 & 79.4 & 78.6 & 78.1 & 77.8 \\
\bottomrule
\end{tabular}
\end{table*}
\begin{table*}[!h]
\centering
\caption{Model performance (accuracy $\%$) by sampling rate}
\label{performance_sr}
\begin{tabular}{lcccccccccccccc}
\toprule
Model & Sampling rate & $2s$ & $4s$ & $8s$ & $16s$ & $32s$ & $64s$ & $128s$ & $256s$ \\
\midrule
\eegmamba$_{\text{temporal}}$ & 250 & 82.4 & 80.5 & 75.8 & 72.5 & 69.4 & 62.6 & 50.3 & 42.9 \\
\eegmamba$_{\text{spectral}}$ & 250 & 77.1 & 76.5 & 75.9 & 75.4 & 74.8 & 74.2 & 73.1 & 72.8 \\
\eegmamba & 250 & 83.1 & 81.6 & 76.7 & 75.7 & 75.4 & 74.6 & 73.4 & 73.0 \\\hline
\eegmamba$_\text{temporal}$ & 125 & 76.7 & 72.5 & 69.8 & 65.6 & 61.7 & 57.3 & 50.4 & 46.7 \\
\eegmamba$_{\text{spectral}}$ & 125 & 69.7 & 68.5 & 67.8 & 67.2 & 66.4 & 64.1 & 62.7 & 60.8 \\
\eegmamba & 125 & 77.0 & 73.6 & 70.3 & 68.5 & 66.7 & 64.7 & 63.1 & 61.2 \\
\bottomrule
\end{tabular}
\end{table*}

\begin{table*}[!h]
\centering
\caption{Model performance (accuracy $\%$) by number of channels}
\label{performance_channel}
\begin{tabular}{lcccccccccccccc}
\toprule
Model & Channels & $2s$ & $4s$ & $8s$ & $16s$ & $32s$ & $64s$ & $128s$ & $256s$ \\
\midrule
\eegmamba$_{\text{temporal}}$ & 6 & 68.4 & 66.7 & 61.2 & 57.4 & 53.2 & 48.4 & 43.1 & 38.8 \\
\eegmamba$_{\text{spectral}}$ & 6 & 60.6 & 59.5 & 58.3 & 57.7 & 56.4 & 55.7 & 54.1 & 53.8 \\
\eegmamba & 6 & 69.2 & 67.5 & 61.3 & 58.1 & 57.4 & 56.8 & 55.1 & 54.2 \\\hline
\eegmamba$_{\text{temporal}}$ & 12 & 76.7 & 74.5 & 72.1 & 68.7 & 63.4 & 61.9 & 52.7 & 44.6 \\
\eegmamba$_{\text{spectral}}$ & 12 & 77.1 & 76.8 & 76.6 & 75.7 & 75.2 & 68.7 & 67.1 & 65.9 \\
\eegmamba & 12 & 78.4 & 77.5 & 77.3 & 76.8 & 75.4 & 69.6 & 67.3 & 66.8 \\
\bottomrule
\end{tabular}
\end{table*}

\paragraph{Benchmark models performance} In assessing the performance of various models for classifying resting state neural disorders, Table \ref{test_accuracy} presents a comparison across different segment lengths ranging from 2s to 256s. The RNN model, with the least number of parameters at 8.6K, shows moderate performance, peaking at 42.7\% accuracy for the 32s segment. EEG-Net, with a slightly higher parameter count of 12K, demonstrates improved accuracy, particularly at 4s with 55.6\%. The EEG-Transformer, despite a significant increase in parameters to 120K, attains its best performance at a shorter 2s segment with 70.1\% accuracy but shows a downward trend as the segment length increases.

\begin{figure}[!ht]
    \centering
    \includegraphics[width=0.45\textwidth]{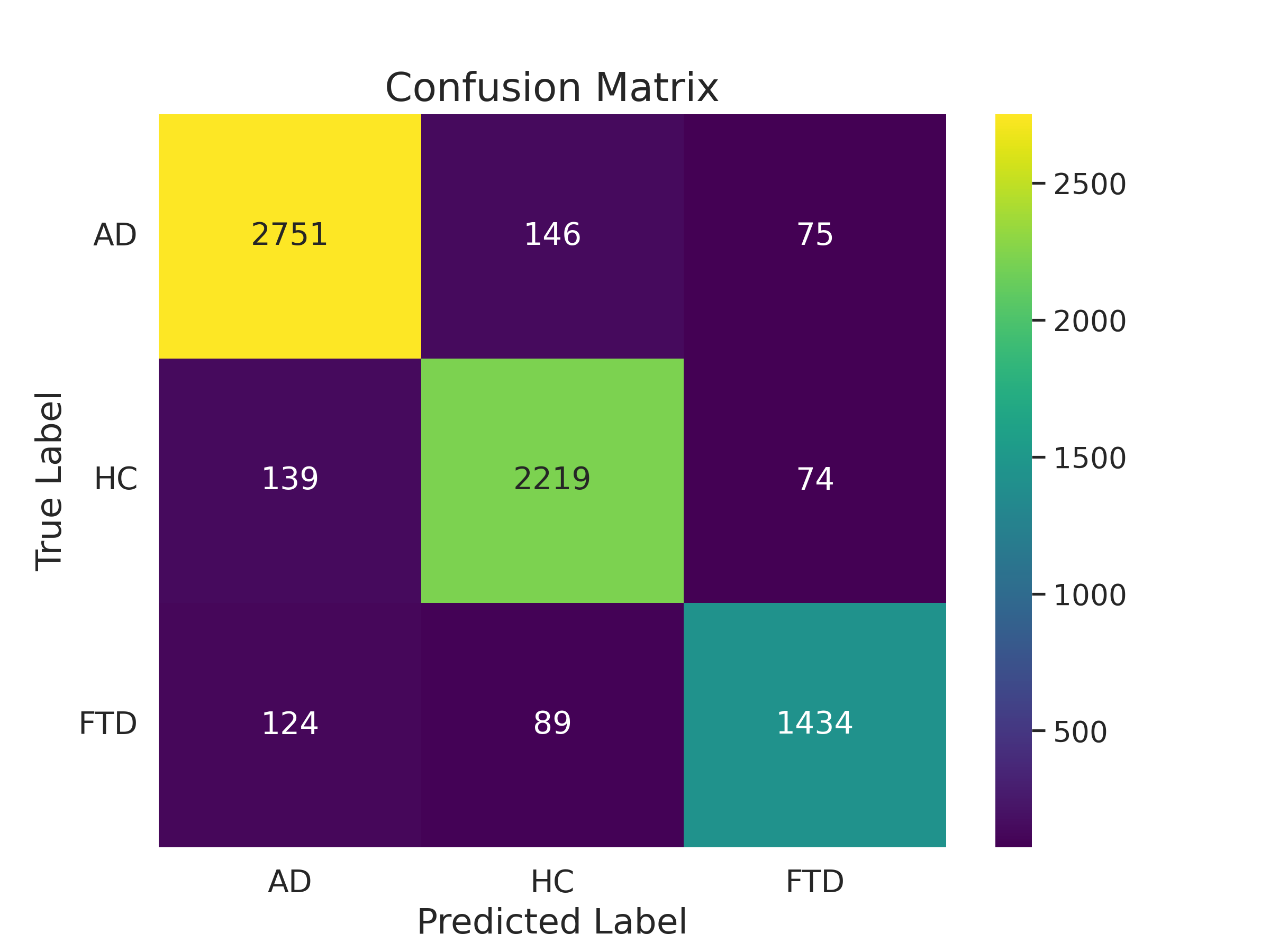}
    \caption{Confusion matrix displaying classification results on the test set using 2-second EEG data segments.}
    \label{confusion_matrix}
\end{figure}

\paragraph{\eegmamba performance} The \eegmamba  model exhibits superior performance across all variants when compared to the RNN and EEG-Net, with the \eegmamba  variant, boasting 4.4K parameters, outperforming other models in longer segments. This variant achieves an impressive 91.0\% accuracy at 2s and maintains robust performance as the segment length increases, highlighting its efficacy in handling various temporal resolutions of EEG data. The \eegmamba$_{\text{temporal}}$ and \eegmamba$_{\text{spectral}}$ variants also show strong results, particularly favoring the mid-range segment lengths of 16s to 64s. However, a common trend across all models is a decline in performance at the longest segment length of 256s, with the  \eegmamba$_{\text{temporal}}$ variant experiencing the most significant drop to 46.8\% accuracy, suggesting a potential trade-off between segment length and model efficacy. The provided confusion matrix, depicted in Figure \ref{confusion_matrix}, corresponds to the performance of the \eegmamba model, which attained an impressive accuracy rate of 91.0\% on a test set derived from 2-second EEG data segments. The total number of 2-second segments for all subjects amounted to 35,254, with a data split ratio of 60\% for training, 20\% for validation, and 20\% for testing. The matrix showcases the model's considerable accuracy in correctly identifying 2,751 instances of AD, 2,219 instances of HC, and 1,434 instances of FTD. Misclassifications were recorded with AD misidentified as HC and FTD in 146 and 75 instances, respectively; HC mistaken for AD and FTD in 139 and 74 instances, respectively; and FTD confounded with AD and HC in 124 and 89 instances, respectively. These figures highlight the model's robust performance in distinguishing AD and HC, indicating a need for further refinement in accurately classifying FTD to enhance its discernment from AD and HC conditions.

\paragraph{The role of EEG sampling rate in \eegmamba model performance} Table \ref{performance_sr} delineates the performance of the \eegmamba model variants to the number of EEG channels at two distinct sampling rates, 250 Hz and 125 Hz. When analyzing the results at the higher sampling rate of 250 Hz, the \eegmamba model stands out with consistently high accuracy across all segment lengths, achieving its peak accuracy of 83.1\% at the 2s segment length. The \eegmamba$_{\text{spectral}}$ variant, while slightly lagging behind its counterparts, still maintains a commendable performance, especially at 256s, where it nearly matches the \eegmamba model with an accuracy of 72.8\%.

The performance trend alters slightly at the lower sampling rate of 125 Hz. Here, all \eegmamba model variants generally show a modest reduction in accuracy across the board. Nevertheless, the \eegmamba$_{\text{combined}}$ model demonstrates resilience, maintaining a strong performance with the highest accuracy of 73.6\% at 4s. Interestingly, the performance gap between the temporal and spectral variants narrows in the longer segment lengths, indicating a potential dependency of model performance on the interplay between sampling rate and EEG data resolution.

Across both sampling rates, there is a notable trend where accuracy tends to decrease with increasing segment length, affirming the complexity of capturing informative features from extended data sequences. The \eegmamba$_{\text{temporal}}$ variant, in particular, shows a significant decline in performance at the longest segment of 256s across both sampling rates, which may suggest limitations in the model's ability to process temporal features over extended periods. Conversely, the robustness of the \eegmamba$_{\text{combined}}$ model across varied segment lengths underscores its effectiveness in integrating multiple feature domains for neural disorder classification, even as the sampling rate varies.

\paragraph{The role of EEG channels in \eegmamba model performance} In Table \ref{performance_channel}, we assess the \eegmamba model's performance across different numbers of EEG channels, specifically examining various combinations of 6 and 12-channel configurations across various segment lengths ranging from 2 seconds to 256 seconds, the highest accuracy of 6 and 12-channels combination is selected to report. The \eegmamba$_{\text{temporal}}$ model, which focuses on temporal features, shows a performance boost when the number of channels is increased from 6 to 12, with accuracy improving from 68.4\% to 76.7\% for the 2s segment length. This trend of improved performance with more channels is consistent across all segment lengths for the temporal model, although the accuracy decrement with longer segments remains evident.

The \eegmamba$_{\text{spectral}}$ model, which relies on spectral features, also benefits from increased channels. With 6 channels, the spectral model peaks at 60.6\% accuracy at the 2s segment, while the 12-channel configuration sees a notable jump to 77.1\% accuracy for the same segment length. The spectral model retains higher accuracy rates with increasing segment lengths when more channels are utilized, suggesting that additional channels provide a richer spectral representation that the model can leverage for classification.

The \eegmamba$_{\text{combined}}$ model, which integrates both temporal and spectral features, achieves the highest accuracy across both channel configurations. For the 6-channel setup, the combined model's best performance is at 4s segment length with 67.5\% accuracy, while the 12-channel configuration sees its peak performance of 78.4\% at the 2s segment length. This model's accuracy is superior at shorter segments and remains relatively stable across longer segments, highlighting the combined model's robustness and ability to utilize a more comprehensive feature set effectively.

The overall trend observed in Table \ref{performance_channel} suggests that the performance of all \eegmamba model variants is influenced by the number of EEG channels and the segment length. There is a clear pattern where the combined model consistently outperforms the individual temporal and spectral models, indicating that a multifaceted approach to feature integration can yield more accurate neural disorder classification across various EEG data resolutions.

\section{Discussion}
In our exploration, we delve into the capabilities of the Mamba model, tailoring it to effectively process multivariate EEG data across a spectrum of temporal resolutions. This endeavor led to creating the \eegmamba temporal variant, highlighting the model's adaptability to diverse EEG segment lengths. Additionally, we introduced the \eegmamba spectral variant, an innovative methodology that facilitates incorporating spectral data from EEG signals into the Mamba framework. The fusion of these two approaches culminates in the \eegmamba combined model, which highlights the advantages of both temporal and spectral analyses to deliver superior accuracy and robustness against variations in EEG segment lengths.
\paragraph{EEG Sequence Length Impact} The Mamba model has demonstrated exceptional performance in natural language processing (NLP) tasks, effectively handling sequence lengths up to \(2^{20}\) tokens \cite{mamba}. However, our findings, as presented in Table \ref{test_accuracy}, reveal a notable performance dip for the \eegmamba$_{\text{temporal}}$ variant as the EEG sequence length increases. This decrease in accuracy can be attributed to the model treating each EEG timepoint as an individual token, where the timepoint token dimension (\(19, 1\)) is significantly smaller compared to typical word token embeddings such as Word2Vec \cite{church2017word2vec}.
\paragraph{EEG Channel Effectiveness}
In our experiments with reduced subsets of EEG channels (specifically, using 6 and 12 channels), we observed a decline in accuracy across all \eegmamba model variants as the number of channels was reduced. This trend likely arises from the unique brain signal information conveyed by each channel in multivariate EEG data, emphasizing the importance of spatial features representing brain connectivity between EEG channels. Despite its success in extracting temporal and spectral features, the current iteration of the \eegmamba model does not incorporate spatial features, marking a limitation and an avenue for future research. The challenge in integrating spatial features lies in their differing data shape compared to temporal and spectral features. Moving forward, we aim to develop methodologies to incorporate spatial features into the \eegmamba model, enhancing its screening capabilities by leveraging the comprehensive neural information in EEG data.
\paragraph{Spectral Feature Effectiveness}
A pivotal aspect of the \eegmamba spectral variant is its capacity to account for individual differences in EEG signal characteristics, thereby optimizing the contribution of distinct EEG frequency bands to each subject's data analysis. This nuanced approach not only underscores the flexibility and computational efficiency of the \eegmamba models but also improves the performance of screening methodologies within clinical neuroscience.  The sequence length issue of Mamba prompted us to explore the incorporation of spectral EEG features as a means to maintain consistent \eegmamba performance across different segment lengths. The PSD extraction from EEG segments offers a promising solution, as variations less influence PSD values in segment length, providing a stable feature set for the \eegmamba model. As evidenced in Table \ref{test_accuracy}, integrating PSD features has resulted in more stable model performance across various EEG segment lengths, underscoring the effectiveness of spectral information in enhancing the \eegmamba framework. Upon evaluating the \eegmamba models across varying sampling rates (Table \ref{performance_sr}), we observed a uniform decline in performance with the reduction of sampling rates. This outcome is intuitively expected, as lower sampling rates yield fewer temporal data points for model training, resulting in diminished EEG data information. Notably, the performance of the \eegmamba$_{\text{temporal}}$ model is particularly sensitive to reductions in sampling rate, a vulnerability that becomes more pronounced with longer segment lengths. This sensitivity underscores the critical role of sampling rate in preserving the integrity of temporal information within the EEG data, with lower rates leading to more significant information loss in the temporal domain.
\paragraph{Temporal and Spectral Combination Effectiveness}
It is crucial to highlight that integrating temporal and spectral features for training with the \eegmamba model results in higher and more consistent accuracy than training the model with unimodal features. We achieved the peak accuracy of 91.0\% when training \eegmamba with 2-second segments. This performance surpasses that of existing studies utilizing the same dataset. For instance, Miltiadous' approach, utilizing a Convolution-Transformer Architecture named DICE-Net \cite{miltiadous2023dice}, reported accuracies of 83.28\% and 74.96\% for AD-CN and FTD-CN models, respectively. Additionally, a method combining CNNs and ViTs detailed in \cite{chen2023multi} achieved an accuracy of 76.37\%. These comparisons underscore the efficacy of the \eegmamba model in dementia detection, highlighting its potential for advancing screening tools in clinical neuroscience.
\section{Conclusion}
Our study presents a novel method for analysing multivariate EEG data through the novel application of the Mamba model, tailored for varying temporal resolutions. The development of \eegmamba temporal and \eegmamba spectral models marks a pivotal step towards leveraging the full spectrum of EEG data, encompassing both temporal and spectral information. This dual approach has demonstrated exceptional performance, notably achieving an accuracy of 91.0\% in distinguishing between HC, FTD, and AD subjects, thereby outperforming existing methodologies on the same dataset. Including \eegmamba spectral is particularly noteworthy for its ability to account for individual variability in EEG signals, ensuring the model's broad applicability and enhanced screening precision. Our findings underscore the potential of \eegmamba models to improve clinical screening, offering a more accurate, efficient, and cost-effective tool for screening to detect dementia early. This research showcases the versatility and effectiveness of the \eegmamba framework. It lays a solid foundation for future research in clinical neuroscience, promising improved screening solutions for a spectrum of neurological disorders.

\bibliographystyle{named}
\bibliography{ijcai19.bib}

\begin{thebibliography}{}

\bibitem[\protect\citeauthoryear{AlSharabi \bgroup \em et al.\egroup }{2022}]{alsharabi2022eeg}
Khalil AlSharabi, Yasser~Bin Salamah, Akram~M Abdurraqeeb, Majid Aljalal, and Fahd~A Alturki.
\newblock Eeg signal processing for alzheimer’s disorders using discrete wavelet transform and machine learning approaches.
\newblock {\em IEEE Access}, 10:89781--89797, 2022.

\bibitem[\protect\citeauthoryear{Association and others}{2018}]{alzheimer20182018}
Alzheimer's Association et~al.
\newblock 2018 alzheimer's disease facts and figures.
\newblock {\em Alzheimer's \& Dementia}, 14(3):367--429, 2018.

\bibitem[\protect\citeauthoryear{Beerens \bgroup \em et al.\egroup }{2014}]{beerens2014quality}
Hanneke~C Beerens, Caroline Sutcliffe, Anna Renom-Guiteras, Maria~E Soto, Riitta Suhonen, Adela Zabalegui, Christina B{\"o}kberg, Kai Saks, Jan~PH Hamers, RightTimePlaceCare Consortium, et~al.
\newblock Quality of life and quality of care for people with dementia receiving long term institutional care or professional home care: the european righttimeplacecare study.
\newblock {\em Journal of the American Medical Directors Association}, 15(1):54--61, 2014.

\bibitem[\protect\citeauthoryear{Borson \bgroup \em et al.\egroup }{2000}]{borson2000mini}
Soo Borson, James Scanlan, Michael Brush, Peter Vitaliano, and Ahmed Dokmak.
\newblock The mini-cog: a cognitive ‘vital signs’ measure for dementia screening in multi-lingual elderly.
\newblock {\em International journal of geriatric psychiatry}, 15(11):1021--1027, 2000.

\bibitem[\protect\citeauthoryear{Bruno and Schurmann~Vignaga}{2019}]{bruno2019addenbrooke}
Diana Bruno and Sofia Schurmann~Vignaga.
\newblock Addenbrooke’s cognitive examination iii in the diagnosis of dementia: a critical review.
\newblock {\em Neuropsychiatric disease and treatment}, pages 441--447, 2019.

\bibitem[\protect\citeauthoryear{Chen \bgroup \em et al.\egroup }{2023a}]{chen2023multi}
Yonglin Chen, Huabin Wang, and Dailei Zhang.
\newblock Multi-feature fusion learning for alzheimer's disease prediction using eeg signals in resting state.
\newblock {\em Frontiers in neuroscience}, 17:1272834, 2023.

\bibitem[\protect\citeauthoryear{Chen \bgroup \em et al.\egroup }{2023b}]{Chen_2023_CVPR}
Zijiao Chen, Jiaxin Qing, Tiange Xiang, Wan~Lin Yue, and Juan~Helen Zhou.
\newblock Seeing beyond the brain: Masked modeling conditioned diffusion model for human vision decoding.
\newblock In {\em Proceedings of the IEEE/CVF Conference on Computer Vision and Pattern Recognition (CVPR)}, 2023.

\bibitem[\protect\citeauthoryear{Chikhi \bgroup \em et al.\egroup }{2022}]{chikhi2022eeg}
Samy Chikhi, Nadine Matton, and Sophie Blanchet.
\newblock Eeg power spectral measures of cognitive workload: A meta-analysis.
\newblock {\em Psychophysiology}, 59(6):e14009, 2022.

\bibitem[\protect\citeauthoryear{Church}{2017}]{church2017word2vec}
Kenneth~Ward Church.
\newblock Word2vec.
\newblock {\em Natural Language Engineering}, 23(1):155--162, 2017.

\bibitem[\protect\citeauthoryear{Delorme \bgroup \em et al.\egroup }{2007}]{delorme2007enhanced}
Arnaud Delorme, Terrence Sejnowski, and Scott Makeig.
\newblock Enhanced detection of artifacts in eeg data using higher-order statistics and independent component analysis.
\newblock {\em Neuroimage}, 34(4):1443--1449, 2007.

\bibitem[\protect\citeauthoryear{Folstein}{1975}]{folstein1975practical}
Marshal~F Folstein.
\newblock A practical method for grading the cognitive state of patients for the clinician.
\newblock {\em J Psychiatr res}, 12:189--198, 1975.

\bibitem[\protect\citeauthoryear{Gu and Dao}{2023}]{mamba}
Albert Gu and Tri Dao.
\newblock Mamba: Linear-time sequence modeling with selective state spaces.
\newblock {\em arXiv preprint arXiv:2312.00752}, 2023.

\bibitem[\protect\citeauthoryear{Guo \bgroup \em et al.\egroup }{2024}]{guo2024mambamorph}
Tao Guo, Yinuo Wang, and Cai Meng.
\newblock Mambamorph: a mamba-based backbone with contrastive feature learning for deformable mr-ct registration.
\newblock {\em arXiv preprint arXiv:2401.13934}, 2024.

\bibitem[\protect\citeauthoryear{Hamilton}{1994}]{hamilton1994state}
James~D Hamilton.
\newblock State-space models.
\newblock {\em Handbook of econometrics}, 4:3039--3080, 1994.

\bibitem[\protect\citeauthoryear{Hogan \bgroup \em et al.\egroup }{2016}]{hogan2016prevalence}
David~B Hogan, Nathalie Jett{\'e}, Kirsten~M Fiest, Jodie~I Roberts, Dawn Pearson, Eric~E Smith, Pamela Roach, Andrew Kirk, Tamara Pringsheim, and Colleen~J Maxwell.
\newblock The prevalence and incidence of frontotemporal dementia: a systematic review.
\newblock {\em Canadian Journal of Neurological Sciences}, 43(S1):S96--S109, 2016.

\bibitem[\protect\citeauthoryear{Ieracitano \bgroup \em et al.\egroup }{2019}]{ieracitano2019convolutional}
Cosimo Ieracitano, Nadia Mammone, Alessia Bramanti, Amir Hussain, and Francesco~C Morabito.
\newblock A convolutional neural network approach for classification of dementia stages based on 2d-spectral representation of eeg recordings.
\newblock {\em Neurocomputing}, 323:96--107, 2019.

\bibitem[\protect\citeauthoryear{Julayanont and Nasreddine}{2017}]{julayanont2017montreal}
Parunyou Julayanont and Ziad~S Nasreddine.
\newblock Montreal cognitive assessment (moca): concept and clinical review.
\newblock {\em Cognitive screening instruments: A practical approach}, pages 139--195, 2017.

\bibitem[\protect\citeauthoryear{Kim \bgroup \em et al.\egroup }{2023}]{kim2023deep}
Min-jae Kim, Young~Chul Youn, and Joonki Paik.
\newblock Deep learning-based eeg analysis to classify normal, mild cognitive impairment, and dementia: Algorithms and dataset.
\newblock {\em NeuroImage}, 272:120054, 2023.

\bibitem[\protect\citeauthoryear{Kingma and Ba}{2014}]{kingma2014adam}
Diederik~P Kingma and Jimmy Ba.
\newblock Adam: A method for stochastic optimization.
\newblock {\em arXiv preprint arXiv:1412.6980}, 2014.

\bibitem[\protect\citeauthoryear{Liu \bgroup \em et al.\egroup }{2024a}]{liu2024swin}
Jiarun Liu, Hao Yang, Hong-Yu Zhou, Yan Xi, Lequan Yu, Yizhou Yu, Yong Liang, Guangming Shi, Shaoting Zhang, Hairong Zheng, et~al.
\newblock Swin-umamba: Mamba-based unet with imagenet-based pretraining.
\newblock {\em arXiv preprint arXiv:2402.03302}, 2024.

\bibitem[\protect\citeauthoryear{Liu \bgroup \em et al.\egroup }{2024b}]{liu2024vmamba}
Yue Liu, Yunjie Tian, Yuzhong Zhao, Hongtian Yu, Lingxi Xie, Yaowei Wang, Qixiang Ye, and Yunfan Liu.
\newblock Vmamba: Visual state space model.
\newblock {\em arXiv preprint arXiv:2401.10166}, 2024.

\bibitem[\protect\citeauthoryear{Ma \bgroup \em et al.\egroup }{2024}]{ma2024u}
Jun Ma, Feifei Li, and Bo~Wang.
\newblock U-mamba: Enhancing long-range dependency for biomedical image segmentation.
\newblock {\em arXiv preprint arXiv:2401.04722}, 2024.

\bibitem[\protect\citeauthoryear{McBride \bgroup \em et al.\egroup }{2014}]{mcbride2014spectral}
Joseph~C McBride, Xiaopeng Zhao, Nancy~B Munro, Charles~D Smith, Gregory~A Jicha, Lee Hively, Lucas~S Broster, Frederick~A Schmitt, Richard~J Kryscio, and Yang Jiang.
\newblock Spectral and complexity analysis of scalp eeg characteristics for mild cognitive impairment and early alzheimer's disease.
\newblock {\em Computer methods and programs in biomedicine}, 114(2):153--163, 2014.

\bibitem[\protect\citeauthoryear{Medsker and Jain}{2001}]{medsker2001recurrent}
Larry~R Medsker and LC~Jain.
\newblock Recurrent neural networks.
\newblock {\em Design and Applications}, 5(64-67):2, 2001.

\bibitem[\protect\citeauthoryear{Miltiadous \bgroup \em et al.\egroup }{2023a}]{miltiadous2023dice}
Andreas Miltiadous, Emmanouil Gionanidis, Katerina~D Tzimourta, Nikolaos Giannakeas, and Alexandros~T Tzallas.
\newblock Dice-net: a novel convolution-transformer architecture for alzheimer detection in eeg signals.
\newblock {\em IEEE Access}, 2023.

\bibitem[\protect\citeauthoryear{Miltiadous \bgroup \em et al.\egroup }{2023b}]{miltiadous2023dataset}
Andreas Miltiadous, Katerina~D Tzimourta, Theodora Afrantou, Panagiotis Ioannidis, Nikolaos Grigoriadis, Dimitrios~G Tsalikakis, Pantelis Angelidis, Markos~G Tsipouras, Euripidis Glavas, Nikolaos Giannakeas, et~al.
\newblock A dataset of scalp eeg recordings of alzheimer’s disease, frontotemporal dementia and healthy subjects from routine eeg.
\newblock {\em Data}, 8(6):95, 2023.

\bibitem[\protect\citeauthoryear{Nichols \bgroup \em et al.\egroup }{2019}]{nichols2019global}
Emma Nichols, Cassandra~EI Szoeke, Stein~Emil Vollset, Nooshin Abbasi, Foad Abd-Allah, Jemal Abdela, Miloud Taki~Eddine Aichour, Rufus~O Akinyemi, Fares Alahdab, Solomon~W Asgedom, et~al.
\newblock Global, regional, and national burden of alzheimer's disease and other dementias, 1990--2016: a systematic analysis for the global burden of disease study 2016.
\newblock {\em The Lancet Neurology}, 18(1):88--106, 2019.

\bibitem[\protect\citeauthoryear{Pohlmann}{2000}]{pohlmann2000principles}
Ken~C Pohlmann.
\newblock {\em Principles of digital audio}.
\newblock McGraw-Hill Professional, 2000.

\bibitem[\protect\citeauthoryear{Rajan \bgroup \em et al.\egroup }{2021}]{rajan2021population}
Kumar~B Rajan, Jennifer Weuve, Lisa~L Barnes, Elizabeth~A McAninch, Robert~S Wilson, and Denis~A Evans.
\newblock Population estimate of people with clinical alzheimer's disease and mild cognitive impairment in the united states (2020--2060).
\newblock {\em Alzheimer's \& dementia}, 17(12):1966--1975, 2021.

\bibitem[\protect\citeauthoryear{Sedghizadeh \bgroup \em et al.\egroup }{2022}]{sedghizadeh2022network}
Mohammad~Javad Sedghizadeh, Hamid Aghajan, Zahra Vahabi, Seyyedeh~Nahaleh Fatemi, and Arshia Afzal.
\newblock Network synchronization deficits caused by dementia and alzheimer’s disease serve as topographical biomarkers: a pilot study.
\newblock {\em Brain Structure and Function}, 227(9):2957--2969, 2022.

\bibitem[\protect\citeauthoryear{Sun \bgroup \em et al.\egroup }{2024}]{sun2024ensemble}
Jingnan Sun, Yike Sun, Anruo Shen, Yunxia Li, Xiaorong Gao, and Bai Lu.
\newblock An ensemble learning model for continuous cognition assessment based on resting-state eeg.
\newblock {\em npj Aging}, 10(1):1, 2024.

\bibitem[\protect\citeauthoryear{Swarnalatha and others}{2023}]{swarnalatha2023greedy}
R~Swarnalatha et~al.
\newblock A greedy optimized intelligent framework for early detection of alzheimer’s disease using eeg signal.
\newblock {\em Computational Intelligence and Neuroscience}, 2023, 2023.

\bibitem[\protect\citeauthoryear{Vaswani \bgroup \em et al.\egroup }{2017}]{vaswani2017attention}
Ashish Vaswani, Noam Shazeer, Niki Parmar, Jakob Uszkoreit, Llion Jones, Aidan~N Gomez, {\L}ukasz Kaiser, and Illia Polosukhin.
\newblock Attention is all you need.
\newblock {\em Advances in neural information processing systems}, 30, 2017.

\bibitem[\protect\citeauthoryear{Welch}{1967}]{welch1967use}
Peter Welch.
\newblock The use of fast fourier transform for the estimation of power spectra: a method based on time averaging over short, modified periodograms.
\newblock {\em IEEE Transactions on audio and electroacoustics}, 15(2):70--73, 1967.

\bibitem[\protect\citeauthoryear{Xu \bgroup \em et al.\egroup }{1994}]{xu1994wavelet}
Yansun Xu, John~B Weaver, Dennis~M Healy, and Jian Lu.
\newblock Wavelet transform domain filters: a spatially selective noise filtration technique.
\newblock {\em IEEE transactions on image processing}, 3(6):747--758, 1994.

\bibitem[\protect\citeauthoryear{Zhu \bgroup \em et al.\egroup }{2024}]{zhu2024vision}
Lianghui Zhu, Bencheng Liao, Qian Zhang, Xinlong Wang, Wenyu Liu, and Xinggang Wang.
\newblock Vision mamba: Efficient visual representation learning with bidirectional state space model.
\newblock {\em arXiv preprint arXiv:2401.09417}, 2024.

\end{thebibliography}

\end{document}